\pdfoutput=1

\documentclass[11pt]{article}

\usepackage[table]{xcolor}

\usepackage[final]{acl}

\usepackage{times}
\usepackage{latexsym}

\usepackage[T1]{fontenc}

\usepackage[utf8]{inputenc}

\usepackage{microtype}

\usepackage{inconsolata}

\usepackage{graphicx}

\usepackage{multirow}
\usepackage{wrapfig}  
\usepackage{pifont}
\usepackage{algorithm}
\usepackage{algorithmic}
\usepackage{amsmath}
\usepackage{subcaption} 
\usepackage{booktabs}
\usepackage{listings}
\usepackage{fancyvrb}
\usepackage{enumitem}
\usepackage{amsthm}
\newtheorem{remark}{Remark}
\definecolor{lightblue}{rgb}{0.9, 0.9, 1}
\lstdefinestyle{mystyle}{
    backgroundcolor=\color{gray!20}, 
    basicstyle=\small\ttfamily,        
    breaklines=true                     
}

\title{VoxEval: Benchmarking the Knowledge Understanding Capabilities of End-to-End Spoken Language Models}

\author{
 \textbf{Wenqian Cui\textsuperscript{1}}\thanks{Corresponding author: wenqian.cui@link.cuhk.edu.hk},
 \textbf{Xiaoqi Jiao\textsuperscript{2}},
 \textbf{Ziqiao Meng\textsuperscript{3}},
 \textbf{Irwin King\textsuperscript{1}},
\\
\small{
 \textsuperscript{1}The Chinese University of Hong Kong,
 \textsuperscript{2}LIGHTSPEED STUDIOS, 
 \textsuperscript{3}National University of Singapore
}
}

\begin{document}
\maketitle
\begin{abstract}
With the rising need for speech-based interaction models, end-to-end Spoken Language Models (SLMs) have emerged as a promising solution. While these models require comprehensive world knowledge for meaningful and reliable human interactions, existing question-answering (QA) benchmarks fall short in evaluating SLMs' knowledge understanding due to their inability to support end-to-end speech evaluation and account for varied input audio conditions. To address these limitations, we present VoxEval, a novel SpeechQA benchmark that assesses SLMs' knowledge understanding through pure speech interactions. Our benchmark 1) uniquely maintains speech format for both inputs and outputs, 2) evaluates model robustness across diverse input audio conditions, and 3) pioneers the assessment of complex tasks like mathematical reasoning in spoken format. Systematic evaluation demonstrates that VoxEval presents significant challenges to current SLMs, revealing their sensitivity to varying audio conditions and highlighting the need to enhance reasoning capabilities in future development. We hope this benchmark could guide the advancement of more sophisticated and reliable SLMs.\footnote{VoxEval dataset is available at: \url{https://github.com/dreamtheater123/VoxEval}}
\end{abstract}

\section{Introduction}
\label{sec:introduction}
The rapid progress in Text-based Large Language Models (TLMs) has revolutionized generative AI, enabling seamless communication between humans and AI via written text. However, natural human interactions often occur through spoken language, making it essential to extend these capabilities to speech. End-to-end Spoken Language Models (SLMs) have emerged as a promising solution, enabling direct speech-based communication with AI \cite{SpeechLMsurvey}. For SLMs to facilitate meaningful and reliable interactions, it is crucial to evaluate their understanding of world knowledge---an essential capability for effective human-AI dialogue. Such evaluations ensure that SLMs can handle complex conversational scenarios and provide accurate, contextually appropriate responses.

\begin{figure}[t]
    \centering
    \includegraphics[width=0.49\textwidth]{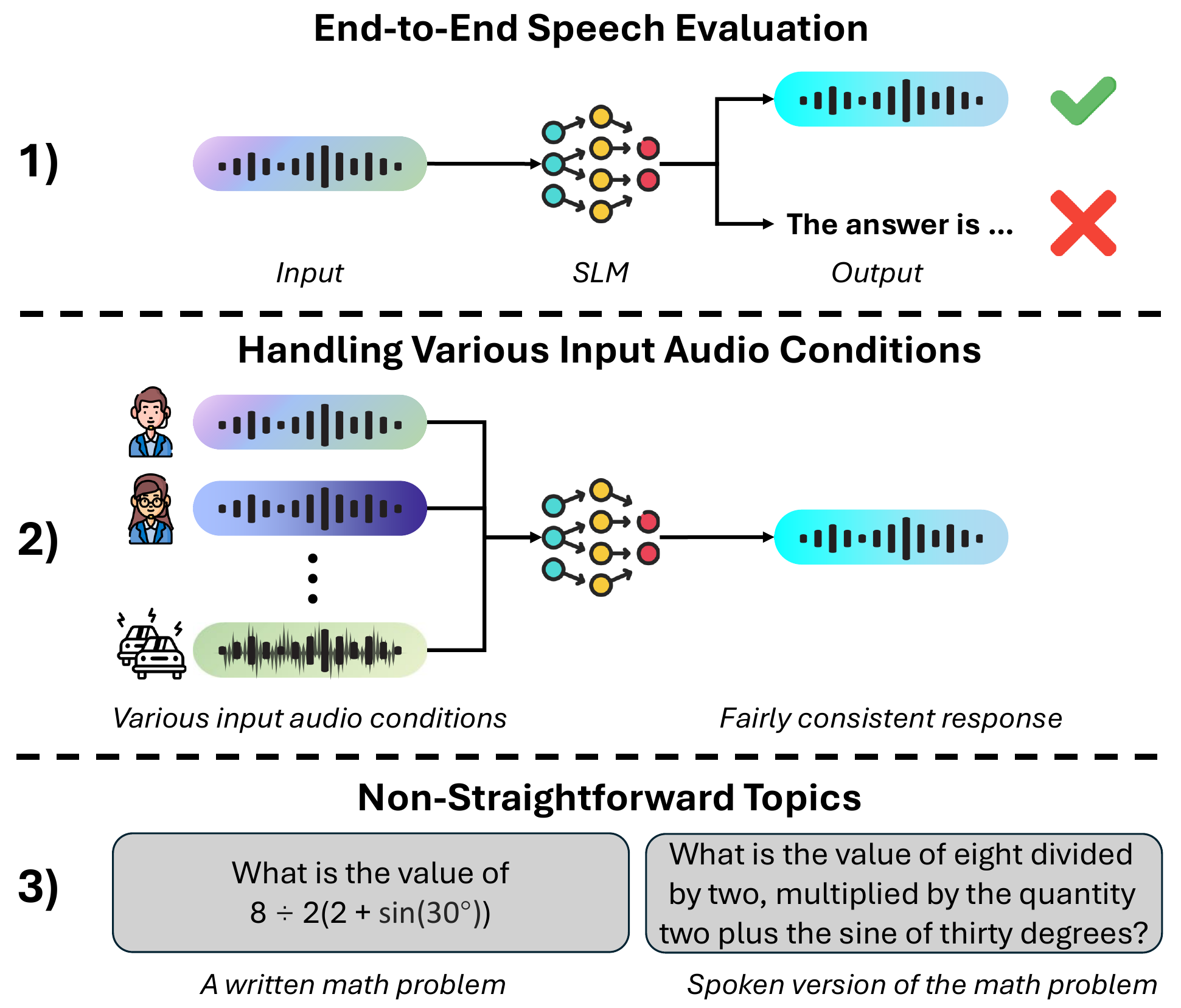} 
    \caption{Illustration of the limitations and challenges of SLM knowledge comprehension evaluation.}
    \label{fig:intro_figure}
\end{figure}

Existing benchmarks \cite{mmau_benchmark,audiobench_benchmark,airbench_benchmark,dynamicsuperb_benchmark} for evaluating audio-based knowledge comprehension face two major limitations. First, they fail to support \textbf{end-to-end speech evaluations}. Current benchmarks construct datasets by either pairing audio clips with text-based QA pairs or providing speech-based questions while evaluating text-based responses. This approach fundamentally misaligns with real-world speech interaction pipelines, where users expect seamless speech-to-speech communication. Evaluating SLMs end-to-end is crucial, as multiple studies have documented substantial performance degradation in SLMs when transitioning from speech-to-text to speech-to-speech evaluations \cite{minmo,vita-audio}. Second, they do not account for \textbf{diverse input audio conditions}, such as variations in speaker timbre, speaking styles, and audio quality, which are common in real-world settings. However, it is crucial for SLMs to demonstrate robustness against these varying conditions, producing consistent answers when the input content remains the same \cite{gpt4o}, especially for fact-based questions. Additionally, we identify a unique challenge in SLM knowledge evaluation: \textbf{not all knowledge comprehension topics are straightforward for SLM evaluation}. For example, mathematical reasoning is easy to evaluate in written form due to the concise nature of mathematical expressions. However, evaluating it in spoken form remains a challenge. Despite this, possessing math reasoning skills is crucial for SLMs, as they are essential for applications like tutoring and simple problem-solving.
Figure \ref{fig:intro_figure} illustrates the three key elements mentioned above.

Based on the key elements outlined, we introduce VoxEval, a SpeechQA benchmark designed to assess the knowledge understanding capabilities of SLMs with the following properties. First, VoxEval includes both questions and answers in audio format, making it, to the best of our knowledge, the \textbf{first benchmark that supports end-to-end evaluation of speech-based interactions}. Second, it incorporates diverse input audio conditions for the same question, allowing us to test the robustness of SLMs across various audio scenarios. Third, to tackle the challenges of evaluating subjects like mathematics and other complex topics, we identify the critical distinctions between written and spoken formats, transform written content into its spoken equivalent, and incorporate Chain-of-Thought prompts to assess the reasoning ability of SLMs in a step-by-step problem-solving setting. Table \ref{tab:ourbenchmarkstatistics} presents the statistics for VoxEval, while Table \ref{tab:benchmarkcomparison} compares the distinctive features of VoxEval with those of other benchmarks. Our evaluation of recently developed SLMs shows that VoxEval poses a substantial challenge for current models, as they perform poorly on VoxEval, exhibit sensitivity to changes in input audio conditions, and demonstrate very limited reasoning capabilities.
To summarize our contributions:
\begin{itemize}[left=0pt]
    \item We introduce a novel benchmark, VoxEval, designed to evaluate the knowledge understanding capabilities of SLMs exclusively in speech format, across a diverse range of audio conditions.
    \item We pioneer the assessment of SLMs' math-solving abilities by introducing a set of spoken math problems and answers.
    \item We conduct a systematic evaluation of VoxEval with several recent SLMs, showing VoxEval is a challenging benchmark to existing SLMs.
\end{itemize}

\section{Related Work}
\label{sec:relatedwork}
\subsection{Speech Large Language Models}
Speech Large Language Models (SpeechLLMs) refers to the LLMs that have the ability to understand and/or generate speech.
There are two primary approaches for integrating speech modalities into LLMs. The first approach focuses on making LLMs understand speech information. This is usually achieved by connecting a speech encoder to a pre-existing LLM while preserving most of the model's original structure and parameters. We refer to these as Speech-to-Text Large Language Models (S2TLLMs). Since the primary goal of these approaches is to enable existing LLMs to perceive auditory information, it only supports text as the output modality \cite{qwen-audio,salmonn}. For example, Qwen-Audio \cite{qwen-audio} utilizes Whisper-large-v2 \cite{whisper} as the audio encoder and Qwen-7B \cite{bai2023qwen} as the LLM. SALMONN \cite{salmonn} uses both whisper and BEATs \cite{chen2022beats} as the audio encoders and uses Vicuna \cite{vicuna2023} to serve as the LLM.

\begin{table}[t]
  \centering
\scalebox{0.94}{
  \begin{tabular}{cccc} 
    \toprule
    \textbf{QA pairs} & \textbf{subjects}  & \textbf{\# IAC} & \textbf{metric} \\
    \midrule
    13,938       & 56 & 26 & ASR + accuracy  \\
    \bottomrule
  \end{tabular}
}
  \caption{\label{tab:ourbenchmarkstatistics}
    Statistics of VoxEval. IAC represents Input Audio Condition.
  }
\end{table}

\begin{table}[t]
  \centering
\scalebox{0.7}{
  \begin{tabular}{lccccc} 
    \toprule
    \textbf{Benchmarks} & \textbf{e2e}  & \textbf{\# KUPs}  & \textbf{\# variations} & \textbf{math} & \textbf{CoT} \\
    \midrule
    SD-Eval       & \textcolor{red}{\ding{55}} & \textcolor{red}{\ding{55}} & \textcolor{red}{\ding{55}} & \textcolor{red}{\ding{55}} & \textcolor{red}{\ding{55}}  \\
    Dynamic-SUPERB       & \textcolor{red}{\ding{55}} & \textcolor{red}{\ding{55}} & \textcolor{red}{\ding{55}} & \textcolor{red}{\ding{55}} & \textcolor{red}{\ding{55}}  \\
    AIR-Bench       & \textcolor{red}{\ding{55}} & \textcolor{red}{\ding{55}} & \textcolor{red}{\ding{55}} & \textcolor{red}{\ding{55}} & \textcolor{red}{\ding{55}}  \\
    AudioBench       & \textcolor{red}{\ding{55}} & 5,196 & \textcolor{red}{\ding{55}} & \textcolor{red}{\ding{55}} & \textcolor{red}{\ding{55}}  \\
    MMAU       & \textcolor{red}{\ding{55}} & $\sim$3,300 & \textcolor{red}{\ding{55}} & \textcolor{red}{\ding{55}} & \textcolor{red}{\ding{55}}  \\
    VoiceBench    & \textcolor{red}{\ding{55}} & 3,074 & 64,695 & \textcolor{red}{\ding{55}} & \textcolor{red}{\ding{55}}  \\
    \textbf{VoxEval}       & \textcolor{green}{\ding{51}} & \textbf{13,938} & \textbf{153,318} & \textcolor{green}{\ding{51}} & \textcolor{green}{\ding{51}}  \\
    \bottomrule
  \end{tabular}
}
  \caption{\label{tab:benchmarkcomparison}
    A comparison of our benchmark with other benchmarks. KUP and CoT stand for Knowledge Understanding Problem and Chain-of-Thought prompts.
  }
\end{table}

In contrast, the second approach advances further by developing end-to-end speech interaction models \cite{speechgpt,spiritlm,defossezmoshi}. In these models, both the input and output modalities are speech, allowing the model to both hear and speak. These approaches directly model speech throughout the LLM-based pipeline and do not significantly rely on textual information. We refer to these models as End-to-End Spoken Language Models (SLMs). SLMs typically involve three components to enable end-to-end speech interactions: tokenizer, language model (LM), and vocoder. Given an input speech, the tokenizer \cite{zhang2023speechtokenizer,hsu2021hubert,w2v-bert} first encodes the audio waveform into speech tokens or representations, LM \cite{llama3,llama2,transformer} then autoregressively generate token response, and vocoder \cite{hifigan,disentangledvocoder,soundstream} synthesizes the output token back into speech waveform.

\subsection{SpeechLLM Evaluation Benchmarks}
The evaluation of SpeechLLMs centers on their ability to model speech data, with researchers examining various aspects and levels. Initially, multiple datasets are utilized to assess the linguistic modeling capabilities of SpeechLLMs. For instance, sWUGGY \cite{speechbenchmark2021} is employed for lexical evaluation, sBLIMP \cite{speechbenchmark2021} for syntactical assessment, and sStoryCloze \cite{textuallypretrainSLM} for semantic evaluation. Additionally, the modeling of paralinguistic elements (spoken content beyond words) is crucial for SpeechLLMs, prompting the development of datasets that evaluate paralinguistic information from both token and perceptual levels \cite{prosodyawareSLM,spiritlm}. Beyond the modeling aspect, the performance of downstream applications is vital for the practical use of SpeechLLMs in real-world scenarios. Recently, numerous downstream evaluation benchmarks have emerged \cite{mmau_benchmark,audiobench_benchmark,airbench_benchmark,dynamicsuperb_benchmark}, focusing on general audio understanding, including speech, environmental sounds, and music. However, as highlighted in Section \ref{sec:introduction}, these benchmarks contain all QA pairs in text format, making them suitable for evaluating S2TLLMs but not SLMs. The closest benchmark to our work is VoiceBench \cite{voicebench}, which evaluates SpeechLLMs under various audio conditions. However, VoiceBench does not support end-to-end evaluation, as it only evaluates based on textual output. Additionally, our study places a stronger emphasis on evaluating the knowledge comprehension abilities of SLMs, not only by including a broader range of topics but also by particularly focusing on their complex skills like mathematics. We also offer a detailed analysis of the performance of SLMs across various scenarios.

\subsection{Knowledge Understanding of LLMs}
\label{sec:relatedworkLLMKnowledgeUnderstanding}
Knowledge understanding is a vital skill for LLMs, as they often need to accurately grasp and produce content based on knowledge. There are many benchmarks to assess the knowledge comprehension of TLMs, each with different levels of difficulty. CommonsenseQA \cite{commonsenseqa} tests commonsense knowledge, which is the most basic level of human understanding. OpenBookQA \cite{openbookqa} focuses on basic science questions. MMLU \cite{MMLU} includes knowledge questions from 57 subjects, mostly at high school or college levels. MMLU-Pro \cite{mmlu-pro} is an advanced version of MMLU with increased difficulty. AGIEval \cite{AGIeval} assesses tasks at a human level, which are also challenging.

\section{VoxEval}
\begin{table*}
  \centering
  \rowcolors{1}{white}{lightblue}
  \small 
  \begin{tabular}{p{3.5cm} p{5.7cm} p{5.5cm}} 
    \toprule
    \textbf{Linguistic Variation}    & \textbf{Explanation} & \textbf{Example} \\
    \midrule
    Filler Words       & Include words like ``um," ``uh," ``like," and ``you know" to see if the model can accurately process the main content despite interruptions.  &  Uh, I’m not sure if I can, like, make it to the meeting on time.     \\
    Mispronunciations     & Introduce common mispronunciations to evaluate the model’s ability to understand intended words.         &  I went to the \textit{libary} this \textit{Febuary} (I went to the library this February)    \\
    Disfluencies       & Test with speech that includes stutters, repetitions, or self-corrections to assess how well the model handles natural speech patterns.           &  It's, it's, it's really important to be, uh, on time. Don't, don't, don't do this again.  \\
    False Starts and Corrections & Insert deliberate errors followed by corrections to assess the model's ability to track changes in speech.     &  What is the value of five times, no, I mean, four times two?   \\
    Language Proficiency    &  Incorporate typical errors made by non-native speakers to evaluate the model's ability to handle non-native grammatical mistakes.   & He just tell me he not coming today.   \\
    \bottomrule
  \end{tabular}
  \caption{\label{tab:linguisticvariation}
    Explanations and examples of the five techniques used in linguistic variation.
  }
\end{table*}

In this section, we introduce a speech-based knowledge understanding benchmark called VoxEval, concentrating primarily on the methods we use to create the data, how we ensure it meets the robustness evaluation criteria, and how we address the evaluation of complex topics like mathematics.


\subsection{Data Construction}
Building a speech-based knowledge understanding benchmark requires the collection of knowledge from different subjects and the construction of QA pairs about the knowledge in speech format. Given that there are numerous existing textual knowledge understanding benchmarks (see Section \ref{sec:relatedworkLLMKnowledgeUnderstanding}), it is logical to build the speech-based benchmark upon textual benchmarks. In this work, we leverage the MMLU \cite{MMLU} test set to build VoxEval. Specifically, we transform the questions from MMLU into speech using the OpenAI Text-to-Speech (TTS) API \footnote{\url{https://platform.openai.com/docs/guides/text-to-speech}}. We select MMLU as the underlying dataset for the following reasons:
\begin{itemize}[left=0pt]
    \item The subjects contained in MMLU are structured and comprehensive, including STEM, social sciences, humanities, and more, and thus ideal for holistic evaluation of knowledge understanding.
    \item MMLU is widely used for evaluating the additional text processing abilities of SLMs \cite{spiritlm,defossezmoshi}. However, this evaluation does not accurately reflect the speech-processing abilities of SLMs.
    \item In datasets available in both written and spoken forms (like sStoryCloze \cite{textuallypretrainSLM}), we find that the spoken version is significantly more challenging than the written one. Therefore, adapting MMLU into a speech format would create a demanding dataset for current SLMs.
\end{itemize}

To transform the MMLU test set into its spoken version, we need to concatenate all the text within a problem into a single sequence. In MMLU, every data is a Multiple Choice Question (MCQ) with four answer choices. We use a simple sentence to concatenate the question and answer choices. Therefore, every converted data looks like this: 
\begin{lstlisting}[style=mystyle]
f"{question} Please choose the answer from options A, B, C, and D. Here are the options. A. {option_A}, B. {option_B}, C. {option_C}, D. {option_D}.", 
\end{lstlisting}
and its corresponding answer is
\begin{lstlisting}[style=mystyle]
f"The correct answer is {answer}.".
\end{lstlisting}

\begin{remark}
Among the 57 MMLU subjects, ``high school computer science" is excluded from VoxEval due to the code snippets within it being unsuitable for verbal evaluation. As a result, VoxEval contains 13,938 unique SpeechQA pairs.
\end{remark}

\subsection{Various Input Conditions}
\label{sec:variousInputConditions}
As mentioned in Section \ref{sec:introduction}, it is crucial for the answers of SLMs to be robust under various input audio conditions. In this work, we define SLM robustness as the capability of an SLM to maintain consistent knowledge understanding QA accuracy when the semantic content of the input remains unchanged but the audio characteristics vary. To evaluate this robustness systematically, we incorporate MMLU questions with different input audio conditions into VoxEval. Specifically, we consider the following types of input conditions.

\subsubsection{Different Speakers}
SLMs should be robust when different speakers interact with them. Several key factors reflect the uniqueness of a speaker, which include but are not limited to gender, age, accent, etc. To evaluate the SLMs' performance on different speakers, we use all six speakers provided by OpenAI TTS, namely alloy, echo, fable, nova, onyx, and shimmer, to perform the TTS. Those speakers span a wide array of speaker properties, such as gender, accent, etc.

\begin{table}
  \centering
  \rowcolors{1}{white}{lightblue}
  \small 
\scalebox{0.955}{
  \begin{tabular}{p{1.7cm} p{1.5cm} p{3.5cm}} 
    \toprule
    \textbf{Circumstance}    & \textbf{Written Version} & \textbf{Spoken Version} \\
    \midrule
    \mbox{Arabic} \mbox{Numerals}       & 2351  &  Two thousand three hundred fifty-one    \\
    Units     & 25cm         &  Twenty-five centimeters    \\
    Operators and Brackets       & $4 \div (2 + 8)$           &  four divided by the sum of two and eight    \\
    \bottomrule
  \end{tabular}
}
  \caption{\label{tab:mathconversion}
    Comparison between written math and spoken math in various circumstances.
  }
\end{table}

\subsubsection{Different Speaking Styles}
In real-world conversations, even a single individual may use different speaking styles depending on the situation. Therefore, SLMs need to manage various input styles. We identify two types of style variations: linguistic and paralinguistic.

\textbf{Linguistic variation} involves changes in the content. For instance, people may sometimes pause to think about what they want to say, leading to disfluencies in their speech. Another example is when someone accidentally says something incorrectly and then corrects themselves upon realizing the mistake. SLMs should still accurately interpret the speech in these situations. However, these scenarios are not typically captured when using TTS to generate speech from a well-written text. To address this, we propose five techniques for linguistic variation to accommodate different situations, including filler words, mispronunciations, disfluencies, false starts and corrections, and language proficiency. Table \ref{tab:linguisticvariation} provides explanations and examples for each of these five variations. To replicate these scenarios, we first use GPT-4o \cite{gpt4o} to modify the original question text into different versions and then convert them into speech. We randomly choose a variation technique for each problem. We only alter the question text and not the answer choice texts. 

\begin{table*}[t]
  \centering
  \small 
\scalebox{1.0}{
  \begin{tabular}{lcccccc} 
    \toprule
    \textbf{Models}  & \textbf{SpeechGPT}  & \textbf{TWIST} & \textbf{SPIRIT-LM} & \textbf{Moshi}  & \textbf{GLM-4-Voice} & \textbf{Whisper + Llama-3.1-8B} \\
    \midrule
    \rowcolor{lightblue} \multicolumn{7}{c}{\textbf{Speakers}} \\ 
    Alloy       & 0.01 & 4.80 & 20.84 & 12.16 & 37.63 & 55.25   \\
    Echo       & 0.01 & 5.58 & 20.96 & 12.21 & 37.64 & 54.68   \\
    Fable       & 0.00 & 1.16 & 20.84 & 11.53 & 36.42 & 53.75   \\
    Nova       & 0.01 & 3.32 & 20.70 & 12.98 & 36.77 & 54.23   \\
    Onyx       & 0.02 & 2.75 & 19.66 & 11.92 & 37.64 & 55.29   \\
    Shimmer       & 0.00 & 5.16 & 20.76 & 12.05 & 38.15 & 55.15   \\
    \midrule
    \rowcolor{lightblue} \multicolumn{7}{c}{\textbf{Speaking Styles}} \\ 
    Linguistic & 0.01 & 4.88 & 20.44 & 11.87 & 36.43 & 55.73   \\
    Speed       & 0.01 & 5.03 & 19.11 & 10.13 & 34.69 & 54.16   \\
    Pitch       & 0.00 & 5.44 & 17.88 & 6.09 & 33.45 & 45.69   \\
    \midrule
    \rowcolor{lightblue} \multicolumn{7}{c}{\textbf{Audio Qualities}} \\ 
    Noise       & 0.00 & 3.68 & 19.50 & 10.18 & 36.95 & 55.51   \\
    Other Env Acou & 0.01 & 4.34 & 20.19 & 10.51 & 37.28 & 55.11 \\
    \bottomrule
  \end{tabular}
}
  \caption{\label{tab:mainresult}
    Evaluation accuracy (\%) of popular SLMs in VoxEval evaluations across different input audio conditions.
  }
\end{table*}

\begin{table}[t]
  \centering
\scalebox{0.85}{
  \begin{tabular}{lcccc} 
    \toprule
    \textbf{Models}  & \textbf{S} $\rightarrow$ \textbf{S}  & \textbf{S} $\rightarrow$ \textbf{T} & \textbf{T} $\rightarrow$ \textbf{S} & \textbf{T} $\rightarrow$ \textbf{T} \\
    \midrule
    SPIRIT-LM       & 20.84 & 23.77 & 1.38 & 17.77  \\
    \bottomrule
  \end{tabular}
}
  \caption{\label{tab:spiritlm4modal}
    Evaluation accuracy (\%) of SPIRIT-LM under different modality combinations. $S$ and $T$ represent speech and text, respectively.
  }
\end{table}

\textbf{Paralinguistic variation}, in contrast, involves changes other than the contents of the input audio. Similar to linguistic variation, SLMs should also output consistent outputs under different paralinguistic variations. We consider two types of paralinguistic variations.
\begin{itemize}[left=0pt]
    \item \textbf{Pitch shift.} Pitch shift refers to the alteration of the pitch of the speech signal, which can simulate the effect of someone speaking in a higher or lower tone. For each question input, we randomly shift its pitch between -5 and 5 semitones.
    \item \textbf{Speed change.} In real-life situations, people often vary their speaking speed, sometimes talking faster and other times slower. To mimic these variations in speaking pace, we apply tempo changes. For each question input, we randomly select a tempo-changing rate that ranges from half the original speed to twice as fast.
\end{itemize}

\subsubsection{Different Audio Qualities}
When interacting with SLMs, users may be in an environment with various acoustic conditions. However, in a standard TTS system, the speech is synthesized in an ``ideal" acoustic environment (very clear, no echo, etc.), thus failing to mimic the real-world interaction scenarios. We consider two types of acoustic conditions when evaluating SLMs:
\begin{itemize}[left=0pt]
    \item \textbf{Background noise.} Sometimes the input audio may be accompanied by background noise, which can affect the clarity and intelligibility of the spoken content. We consider different kinds of noise, including Gaussian noise, colored noise, background music, and short interruption noise. For each problem, a kind of noise is randomly selected and applied to the original audio. Table \ref{tab:noiseparameters} presents the transformation settings for all the noise types used.
    \item \textbf{Other environment acoustics.} We consider speech recorded under various conditions, including aliasing, room impulse response, low pass filter, high pass filter, band pass filter, bitcrush, gain, clipping distortion, and seven band parameter EQ. Full explanations and transformation settings are presented in Appendix \ref{apdx:recordingconditions}. A random audio effect is applied to each question input.
\end{itemize}


\begin{remark}
We apply all the variations to the speech data generated only using the speaker ``alloy". To create different paralinguistic and acoustic conditions, we utilize the \mbox{\textit{``audioaugmentations"}} package\footnote{\url{https://github.com/iver56/audiomentations}} to augment the audio.
\end{remark}



\begin{table*}[t]
  \centering
\scalebox{0.67}{
  \begin{tabular}{lccccc} 
    \toprule
    \textbf{Models (MGS)}  & \textbf{SpeechGPT}  & \textbf{TWIST} & \textbf{SPIRIT-LM} & \textbf{Moshi} & \textbf{GLM-4-Voice} \\
    \midrule
    \rowcolor{lightblue} \multicolumn{6}{c}{\textbf{Speakers}} \\ 
    Alloy       & 0.00/0.00/0.00/\textbf{0.03} & 5.61/3.81/3.19/\textbf{5.87} & 22.44/21.71/19.29/\textbf{23.30} & \textbf{13.09}/11.95/11.98/12.85 & 32.50/38.58/39.73/\textbf{43.03}    \\
    Echo       & \textbf{0.04}/0.00/0.00/0.00 & 6.11/5.48/3.41/\textbf{6.92} & 22.48/21.82/18.67/\textbf{24.22} & \textbf{14.13}/12.46/11.29/13.71 & 33.25/\textbf{42.26}/39.06/41.23    \\
    Fable       & 0.00/0.00/0.00/0.00 & 0.88/0.56/0.44/\textbf{1.70} & 21.22/22.90/19.15/\textbf{23.64} & \textbf{12.57}/11.65/10.42/11.91 & 30.81/\textbf{42.52}/38.32/39.33    \\
    Nova       & 0.00/\textbf{0.02}/0.00/0.01 & 3.49/2.74/1.97/\textbf{4.08} & \textbf{23.02}/21.72/17.99/22.91 & \textbf{15.04}/12.69/10.84/14.32 & 31.95/\textbf{42.80}/38.43/40.12    \\
    Onyx       & \textbf{0.01}/0.00/0.01/0.00 & 3.46/2.73/1.49/\textbf{3.95} & 20.85/21.09/18.23/\textbf{22.42} & \textbf{13.56}/12.74/11.05/13.39 & 32.38/\textbf{44.27}/40.84/38.32    \\
    Shimmer       & 0.00/0.00/0.00/0.00 & 6.86/4.79/3.01/\textbf{6.91} & 21.90/\textbf{22.29}/19.09/21.72 & \textbf{13.87}/12.10/10.85/13.06 & 31.79/\textbf{45.96}/39.35/44.34    \\
    \midrule
    \rowcolor{lightblue} \multicolumn{6}{c}{\textbf{Speaking Styles}} \\ 
    Linguistic & 0.00/0.00/\textbf{0.04}/0.00 & 5.81/4.31/3.05/\textbf{5.88} & 20.62/21.57/20.20/\textbf{22.57} & 13.10/11.67/10.09/\textbf{13.44} & 32.18/38.18/36.94/\textbf{39.91}    \\
    Speed       & 0.00/0.00/0.02/\textbf{0.05} & 5.71/5.14/3.42/\textbf{6.67} & 20.63/\textbf{20.94}/17.83/20.38 & 10.59/11.37/09.16/\textbf{11.71} & 30.55/\textbf{38.74}/36.30/37.98    \\
    Pitch       & 0.00/0.00/0.00/0.00 & 6.20/5.55/3.67/\textbf{6.85} & 17.94/18.44/17.60/\textbf{20.19} & \textbf{06.60}/05.73/05.67/05.85 & 30.03/\textbf{37.35}/34.76/36.43    \\
    \midrule
    \rowcolor{lightblue} \multicolumn{6}{c}{\textbf{Audio Qualities}} \\ 
    Noise       & 0.00/0.00/0.00/0.00 & 4.22/3.37/2.47/\textbf{4.84} & 20.77/\textbf{21.73}/16.59/21.49 & 10.73/\textbf{10.79}/9.83/10.78 & 31.45/\textbf{41.93}/38.68/41.46    \\
    Other Env Acoustics & 0.00/0.00/0.00/\textbf{0.03} & 5.19/3.71/2.79/\textbf{5.46} & 22.44/20.66/16.84/\textbf{23.25} & 11.18/10.08/10.24/\textbf{11.30} & 30.45/\textbf{43.40}/39.09/41.42    \\
    \bottomrule
  \end{tabular}
}
  \caption{\label{tab:mean_max_diff_results}
    Average accuracy (\%) for various subject groups across different input audio conditions are presented. The scores are shown in the format ``STEM/Social Science/Humanities/Others" subjects. MGS stands for Mean Grouped Score. The highest score in each scenario is highlighted in bold.
  }
\end{table*}

\subsection{Handle Math Expressions and More}
Logical reasoning and critical thinking are crucial for AI assistants like LLMs, and tackling math problems is a major part of these abilities. Math-solving skills are vital for SLMs because humans do not always express mathematical ideas in written form, and spoken math has many uses for AI assistants, such as in tutoring. Hence, we pioneer the evaluation of math-solving abilities for SLMs. 

Mathematical problems often include numerous mathematical expressions, and current TTS systems struggle to accurately convert these expressions into speech, as they are typically trained only on textual words. To tackle this issue, we propose a two-step method for synthesizing math problems. First, we convert the written math problems into their spoken form, and then we use the TTS system to generate speech. Specifically, we create a set of few-shot examples and use GPT-4o to handle the conversion process. Our focus is on transforming Arabic numerals, units, operators, and brackets, as illustrated by the examples in Table \ref{tab:mathconversion}. The complete prompt is provided in Figure \ref{fig:questionprompt} and \ref{fig:answerchoiceprompt}.


\begin{figure*}[t]
    \centering
    \subfloat[Box plots showing the absolute score differences using ``alloy" as the reference.]{
        \label{fig:abs_diff_allsubjects}
        \includegraphics[width=0.99\textwidth]{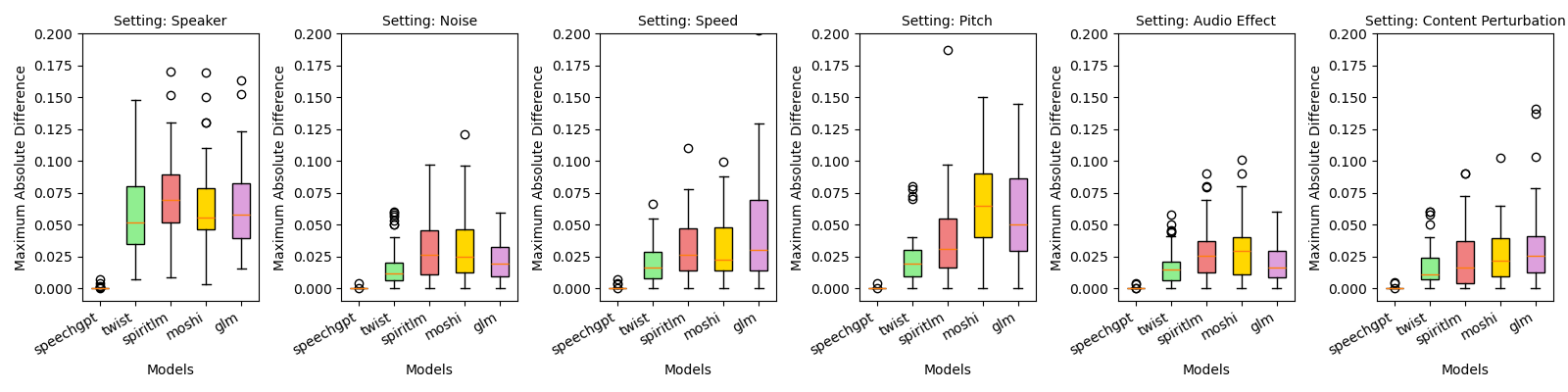}
    }
    \\
    \subfloat[Box plots showing the relative score differences using ``alloy" as the reference.]{
        \label{fig:rel_diff_allsubjects}    
        \includegraphics[width=0.99\textwidth]{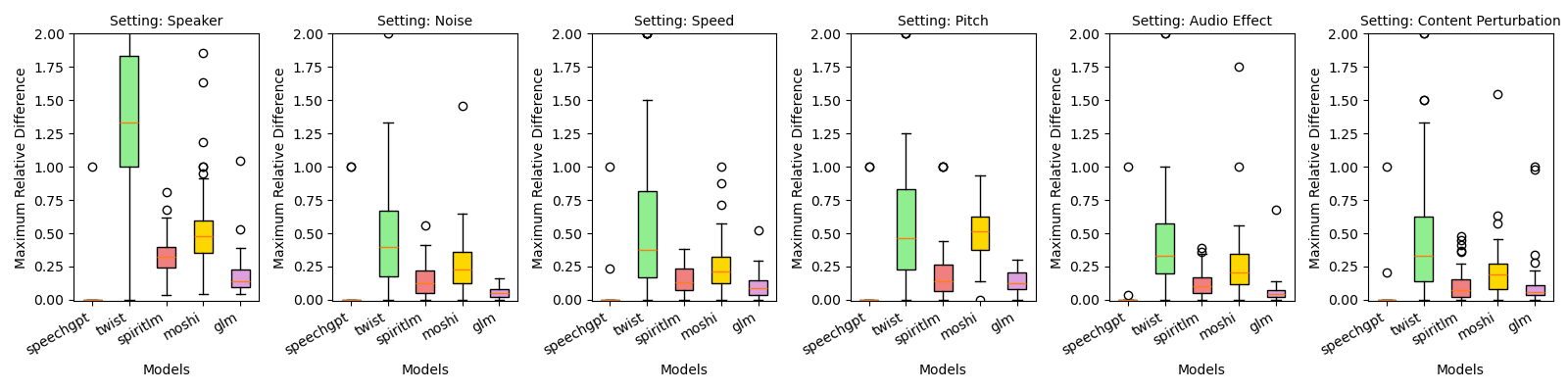}
    }
    \caption{Box plots to display the maximum performance score differences across different settings.}
    \label{fig:diff_allsubjects}
\end{figure*}

\begin{table}[t]
  \centering
  \scriptsize
\scalebox{0.9}{
  \begin{tabular}{l|cc|cc|cc} 
    \toprule
    & \multicolumn{2}{c|}{\textbf{Elementary Math}} & \multicolumn{2}{c|}{\textbf{High School Math}} & \multicolumn{2}{c}{\textbf{College Math}} \\
    \cmidrule(lr){2-3} \cmidrule(lr){4-5} \cmidrule(lr){6-7}
    \textbf{Models} & \textit{Direct} & \textit{LLM} & \textit{Direct} & \textit{LLM} & \textit{Direct} & \textit{LLM} \\
    \midrule
    \rowcolor{lightblue} \multicolumn{7}{c}{\textbf{Direct Answer}} \\ 
    SpeechGPT    & 0.00  & 0.00  & 0.00  & 0.00  & 0.00  & 0.00  \\ 
    TWIST        & 3.17  & 1.59  & 4.44  & 1.11  & 8.00  & 4.00  \\
    SPIRIT-LM    & 23.02 & 20.63 & 20.00 & 20.00 & 22.00 & 25.00 \\
    Moshi        & 11.38 & 11.90 & 19.26 & 18.52 & 17.00 & 14.00 \\
    GLM-4-Voice  & 25.66 & 25.66 & 25.93 & 25.93 & 27.00 & 26.00 \\
    \midrule
    \rowcolor{lightblue} \multicolumn{7}{c}{\textbf{Chain-of-Thought (Cut)}} \\ 
    SpeechGPT    & 0.00  & 0.00  & 0.00  & 0.00  & 0.00  & 0.00  \\ 
    TWIST        & 2.65  & 1.32  & 4.81  & 3.33  & 4.00  & 2.00  \\
    SPIRIT-LM    & 15.87 & 12.17 & 18.51 & 17.41 & 18.00 & 18.00 \\
    Moshi        & 15.08 & 16.40 & 12.59 & 13.70 & 15.00 & 18.00 \\
    GLM-4-Voice  & 21.96 & 23.02 & 5.19 & 4.81 & 7.00 & 5.00 \\
    \midrule
    \rowcolor{lightblue} \multicolumn{7}{c}{\textbf{Chain-of-Thought (No cut)}} \\ 
    SpeechGPT    & 0.00  & 0.00  & 0.00  & 0.00  & 0.00  & 0.00  \\ 
    TWIST        & 0.00  & 0.00  & 0.00  & 0.00  & 0.00  & 0.00  \\ 
    SPIRIT-LM    & 17.20 & 15.34 & 13.33 & 11.48 & 15.00 & 13.00 \\
    Moshi        & 16.67 & 16.67 & 12.96 & 13.33 & 14.00 & 12.00 \\
    GLM-4-Voice  & 23.02 & 22.49 & 7.78 & 7.41 & 6.00 & 5.00 \\
    \bottomrule
    \noalign{\vskip -\arrayrulewidth}
  \end{tabular}
}
  \caption{\label{tab:mathresult}
    Evaluation accuracy (\%) for different levels of math problems under the ``Alloy" speaker setting.
  }
\end{table}

\section{Experiments}
We evaluate recently proposed SLMs on our proposed VoxEval benchmark. In this section, we explain the experimental settings and the corresponding results.


\subsection{Experimental Setups}
\textbf{Data Preparations.} We assess the performance of the SLM across various input audio conditions, including different speakers, speaking styles, and audio qualities, as mentioned in Section \ref{sec:variousInputConditions}. Few-shot prompting techniques are employed to guide the SLMs in answering questions from VoxEval. Specifically, for each subject, we synthesize five examples from the MMLU validation set using all six speaker voices provided by OpenAI TTS. The five in-context examples of the corresponding speaker are prepended to the final question. To accommodate the input constraints of certain SLMs, the resulting speech is truncated to the last 80 seconds.

\textbf{Models.} We evaluate VoxEval on five recently introduced (open-source) SLMs: SpeechGPT \cite{speechgpt}, TWIST \cite{textuallypretrainSLM}, SPIRIT-LM \cite{spiritlm}, Moshi \cite{defossezmoshi}, and GLM-4-Voice \cite{GLM-4-Voice}. These models represent diverse training approaches within the SLM domain. Specifically, SpeechGPT and TWIST are pre-trained using unsupervised speech data. SPIRIT-LM and GLM-4-Voice are pre-trained on interleaved speech and text data. Moshi is pre-trained on aligned speech-text data. All the models are built upon TLM checkpoints. Furthermore, while the other models process audio data through a single stream, Moshi uniquely represents audio data using multiple streams.

\textbf{Evaluation Metric.} To assess the spoken responses provided by the SLMs, we utilize the \mbox{OpenAI} ASR model whisper-large-v3 \cite{whisper} to convert their answers into text. \mbox{Afterward}, we apply \textit{string matching} to determine the final answer (e.g., A, B, C, or D) from the transcription and calculate the accuracy.

\subsection{Results}
We aim to answer the following Research Questions (RQs) through experiments. \textbf{RQ1.} What is the overall performance of existing SLMs on the VoxEval benchmark? \textbf{RQ2.} How do existing SLMs perform across various knowledge domains? \textbf{RQ3.} To what extent do different input audio conditions influence the performance of existing SLMs? \textbf{RQ4.} How well are current SLMs in reasoning, particularly in mathematical reasoning?


To answer \textbf{RQ1}, we evaluate five SLMs on VoxEval, as shown in Table \ref{tab:mainresult}. Results reveal that current SLMs perform \textbf{poorly}, with \textbf{most failing to surpass random guessing}. Since VoxEval questions have four answer choices, random guessing yields an expected score of 25\%. However, only GLM-4-Voice exceeds this baseline, indicating that SLMs often struggle to follow instructions and select the correct answer. This underscores the challenge of enabling SLMs to ``speak" accurate responses, making VoxEval a tough benchmark. A case study of SLMs' spoken responses is in Appendix \ref{apdx:SLMresponses}. Notably, SpeechGPT's performance is near 0\%. This is likely due to our approach of directly performing speech-to-speech generation without utilizing the ``chain-of-modality" mode described in \cite{speechgpt}. We choose not to use the ``chain-of-modality" approach because it involves converting the input into text before generating a response, which we believe would significantly increase inference latency and is not an optimal strategy for end-to-end SLMs.

Since some SLMs support both speech and text as input/output (I/O), we further evaluate the performance of SLM when using the four different I/O modality combinations. In particular, we focus on SPIRIT-LM, as it is the only model that supports both modalities for input and output during inference. The evaluation results, shown in Table \ref{tab:spiritlm4modal}, reveal that SPIRIT-LM's performance varies significantly depending on the modality combination used, with differences exceeding 20\%. This highlights the difficulty of achieving consistent behavior across various I/O configurations.

To answer \textbf{RQ2}, we present the SLM performance results across various subject groups within VoxEval. Following MMLU, all knowledge subjects are divided into four groups: \textit{STEM} (18 subjects), \textit{social science} (12 subjects), \textit{humanities} (13 subjects), and \textit{others} (13 subjects). Table \ref{tab:mean_max_diff_results} shows the average performance of the subjects in each group. We notice that \textbf{SLMs exhibit significant performance variations across different subject groups}. For instance, when using alloy as the input voice, SPIRIT-LM shows a score difference of approximately 4\%, and GLM-4-Voice demonstrates a difference of around 10\%. Moreover, \textbf{different SLMs excel in different subject groups}. For example, despite the change in input audio conditions in various settings, TWIST, Moshi, and GLM-4-Voice typically perform the best in the \textit{others}, \textit{STEM}, and \textit{social science} groups, respectively.

To answer \textbf{RQ3}, we present the overall performance of SLMs across various input audio conditions in Table \ref{tab:mainresult}. Our key findings include:
\begin{itemize}[left=0pt]
\item \textbf{SLMs are susceptible to different input audio conditions.} Although SLM performances do not vary significantly across different speaker voices, variations in speaking styles and audio quality can cause up to a 6\% performance drop.
\item \textbf{Different input audio conditions have different impacts on SLMs.} For instance, environment acoustics and linguistic variations have minimal impact on SLMs' performance, whereas pitch shifts pose the greatest challenge for most SLMs.
\end{itemize}

We further visualize performance differences across all input audio condition settings, as shown in Figure \ref{fig:diff_allsubjects}. For each setting, we compute the maximum score difference for each of the 56 subjects in the VoxEval dataset, based on the scores obtained across all results in the corresponding setting. This yields 56 difference scores for each setting, which are presented using box plots. In the ``Speaker" setting, the differences are calculated using audio from all six speakers provided by OpenAI TTS. For the remaining settings, since all audio variations are applied to speech data generated by the ``alloy" voice, the differences are computed by comparing the scores of the original ``alloy" audio with those of the transformed audio. The top part of the figure illustrates the absolute differences in scores, while the bottom part shows the relative percentage differences. Algorithm \ref{alg:boxplot} presents the pseudo-code for generating the box plots. We make the following key observations.

\begin{itemize}[left=0pt]
    \item \textbf{Performance variation among individual subjects is significantly greater than the overall performance variation.} While overall performance differences are typically within 1–2\%, we found many instances where the performance gap for a specific subject exceeded 10\%.
    \item \textbf{Models exhibit varying levels of stability when faced with changes in audio conditions.} Lower score differences indicate better stability. As shown in Figure \ref{fig:rel_diff_allsubjects}, TWIST and Moshi tend to be less stable, whereas GLM-4-Voice demonstrates minimal relative score differences.
\end{itemize}

To answer \textbf{RQ4}, we present the performance of SLMs across three math subjects in VoxEval, as shown in Table \ref{tab:mathresult}. The selected subjects---elementary math, high school math, and college math---are intended to represent varying levels of reasoning complexity. The math-solving skills of SLMs are evaluated in two ways: direct answer, where the model performs 5-shot inference to provide the answer directly without showing intermediate reasoning, and Chain-of-Thought prompting (CoT) \cite{chainofthoughtCoT}, which uses annotated intermediate steps from three dev-set questions per subject as examples appended before the actual question. All the CoT examples and questions are in speech format. CoT evaluations are conducted in two settings: \textit{Cut}, where the input audio is trimmed to the last 80 seconds (consistent with other experiments), and \textit{No cut}, where the audio is left untrimmed to preserve longer CoT prompts. Two metrics are used for evaluation: \textit{string matching}, the standard final answer extraction method, and \textit{LLM}, where GPT-4o \cite{gpt4o} is used to extract answers to complement string matching, as models often include lengthy reasoning steps that hinder accurate string-based extraction when applying CoT. Our observations include:
\begin{itemize}[left=0pt]
    \item \textbf{Existing SLMs perform similarly across different levels of math questions.} Interestingly, in some cases, these models occasionally achieve better results on more complex problems than on simpler ones. Among them, GLM-4-Voice stands out as the only SLM consistently excelling in simple (elementary) math. This counterintuitive fact suggests that \textbf{most SLMs are unlikely to possess basic math reasoning abilities.}
    \item \textbf{Using CoT reduces the performance of SLMs.} We observe this trend consistently across all tested models. This suggests that CoT does not enhance SLMs in the same way it improves TLMs, emphasizing the need for specialized techniques to improve SLM reasoning.
    \item \textbf{Extending CoT example length does not significantly impact SLM performances.} Most models perform similarly in \textit{Cut} and \textit{No cut} scenarios, but TWIST struggles in \textit{No cut}, suggesting some models are sensitive to input audio length.
\end{itemize}



\section{Conclusion}
This paper introduces VoxEval, a speech question-answering benchmark for assessing the knowledge understanding of end-to-end Spoken Language Models (SLMs) across diverse audio conditions. Evaluations of recent SLMs reveal significant limitations, including low accuracies that are often below random guessing, sensitivity to variations in input audio conditions, and difficulties with reasoning tasks like mathematical problem-solving. These results emphasize the need for further research to improve SLMs' knowledge understanding, robustness, and reasoning capabilities in real-world conversational settings. 
VoxEval serves as a critical step toward advancing the evaluation and development of SLMs, paving the way for more effective and reliable speech-based AI systems.


\section{Limitations}
The limitations of this work come from the scope and the data construction methodology. First, the VoxEval benchmark focuses exclusively on assessing the robustness of SLMs in terms of knowledge understanding, leaving out other critical aspects such as fairness, toxicity, hallucination, etc. These issues are particularly relevant in generative scenarios, where SLMs may exhibit biases, generate harmful content, or produce inaccurate information. However, these aspects fall outside the scope of VoxEval and require separate evaluation frameworks tailored to address them comprehensively. Second, the benchmark's reliance on synthetic speech generation through TTS systems may not fully capture the complexity and variability of natural human speech, including emotional expressions, regional dialects, and spontaneous speech patterns that are difficult to replicate artificially.

\section*{Acknowledgments}
The research presented in this paper was partially funded by the Research Grants Council of the Hong Kong Special Administrative Region, China (CUHK 2410072, RGC R1015-23).

\bibliography{acl_latex}

\appendix

\section{Accuracy of the Answer Extraction Pipeline}
\begin{table}[t]
  \centering
\scalebox{0.72}{
  \begin{tabular}{lccc} 
    \toprule
    \textbf{Models}  & \textbf{GLM-4-Voice}  & \textbf{Moshi} & \textbf{SPIRIT-LM} \\
    \midrule
    WER       & 0.01 & 0.16 & 0.08  \\
    CER       & 0.01 & 0.10 & 0.04  \\
    Total Word Errors       & 3 & 12 & 23  \\
    Total Character Errors       & 13 & 22 & 64  \\
    Total Words in GT       & 300 & 78 & 308  \\
    Total Characters in GT       & 1759 & 401 & 1767  \\
    \bottomrule
  \end{tabular}
}
  \caption{\label{tab:asrevaluation}
    ASR process evaluation results. GT represents Ground Truth.
  }
\end{table}

To ensure the reliability of our automated pipeline (ASR + string matching) for extracting final answers from SLMs' speech output, it is essential to assess its accuracy. This section provides the evaluation results, focusing on two key questions. 1) How accurate is the ASR process in transcribing the SLMs' spoken output into text? 2) Is the final answer extracted correctly?

To answer the above two questions, we randomly select ten questions from the VoxEval benchmark and conduct spoken QA using three top-performing SLMs (i.e., GLM-4-Voice, Moshi, and SPIRIT-LM). This process results in 30 SLM-generated responses. For the first question, we use Whisper-large-v3 to transcribe these 30 responses and also manually transcribe them to serve as ground truth. Finally, we calculate the Word Error Rate (WER) and Character Error Rate (CER) by comparing the ASR transcriptions with the manual ground truth. As shown in Table \ref{tab:asrevaluation}, all three models have very low WER and CER scores. While Moshi exhibits relatively higher error rates, this can be attributed to its characteristically shorter responses compared to the other models. Since WER and CER are normalized by the ground truth length, direct comparison may not fully reflect relative performance. Nevertheless, these results indicate minimal ASR transcription errors across all models. For the second question, using the same 30 model responses, we apply our automatic evaluation pipeline to extract final answers and compare these against manually extracted ground truth answers obtained through direct listening. The accuracy achieved is 29/30. The single error occurred with a SPIRIT-LM response characterized by rapid and unclear speech, resulting in ASR failure. However, this error is attributable to the model's failure to provide a clear, intelligible answer rather than a systematic pipeline limitation. To summarize, these experiments demonstrate the robustness and reliability of our automatic evaluation pipeline, with high accuracy in both ASR transcription and the answer extraction task.

\section{Comparison between Different Speech Interaction Models}
\label{sec:comparisonspeechinteractionmodels}

\begin{table}[t]
  \centering
\scalebox{1.0}{
  \begin{tabular}{lc} 
    \toprule
    \textbf{Models}  & \textbf{Accuracy} \\
    \midrule
    GLM-4-Voice       & 0.472  \\
    Whisper + Llama-3.1       & 0.606  \\
    Whisper + Qwen2       & 0.517  \\
    gpt-4o-audio-preview       & 0.718  \\
    \bottomrule
  \end{tabular}
}
  \caption{\label{tab:differentSpeechInteractionModels}
    Performance comparison between different speech interaction models.
  }
\end{table}

To better understand how well existing end-to-end SLMs are, we compare their performance with other types of speech interaction models. Specifically, we examine two alternative model categories:
\begin{enumerate}[left=0pt]
    \item Open-source cascaded models: These models follow a pipeline structure that combines ASR with a TLM, so they are not considered end-to-end SLMs. This comparison helps determine whether cascaded models or SLMs perform better. For this category, we use two combinations: Whisper-large-v3 + Llama-3.1-8B-Instruct and Whisper-large-v3 + Qwen2-7B-Instruct.
    \item Closed-source API models: These are speech interaction models whose checkpoints are not publicly available and can only be accessed through their official APIs. This comparison aims to identify performance gaps between open-source and closed-source models. For this category, we use gpt-4o-audio-preview.
\end{enumerate}
We include only GLM-4-Voice for SLMs, as it performs the best in our main experiments. Due to budget constraints, we choose three subjects from each of the four subject groups (STEM, social science, humanities, others) of VoxEval, resulting in 12 subjects. For each subject, we keep the first 100 QA pairs for this evaluation.

We observe two key findings from the results shown in Table \ref{tab:differentSpeechInteractionModels}. First, cascaded models outperform SLMs, highlighting unique challenges associated with developing end-to-end models. Second, the closed-source API model achieves the highest performance and demonstrates a notable gap compared to SLMs, suggesting the potential performance ceiling for SLMs is far from being reached.

\section{Further Analysis of the In-context Learning Capabilities of SLMs with and without CoT}
\begin{table}[t]
  \centering
  \scriptsize
\scalebox{0.94}{
  \begin{tabular}{l|cc|cc|cc} 
    \toprule
    & \multicolumn{2}{c|}{\textbf{Elementary Math}} & \multicolumn{2}{c|}{\textbf{High School Math}} & \multicolumn{2}{c}{\textbf{College Math}} \\
    \cmidrule(lr){2-3} \cmidrule(lr){4-5} \cmidrule(lr){6-7}
    \textbf{Models} & \textit{Non-CoT} & \textit{CoT} & \textit{Non-CoT} & \textit{CoT} & \textit{Non-CoT} & \textit{CoT} \\
    \midrule
    0-shot    & 0.07  & 0.07  & 0.06  & 0.06  & 0.02  & 0.02  \\
    1-shot    & 0.25  & 0.28  & 0.20  & 0.04  & 0.25  & 0.08  \\
    2-shot    & 0.24  & 0.31  & 0.21  & 0.05  & 0.27  & 0.11  \\
    3-shot    & 0.26  & 0.28  & 0.26  & 0.08  & 0.27  & 0.08  \\
    4-shot    & 0.25  & -  & 0.27  & -  & 0.25  & -  \\
    5-shot    & 0.23  & -  & 0.26  & -  & 0.31  & -  \\
    \bottomrule
    \noalign{\vskip -\arrayrulewidth}
  \end{tabular}
}
  \caption{\label{tab:morecotresults}
    In-context learning results of GLM-4-Voice with and without CoT. Non-CoT represents in-context learning without CoT (direct answer).
  }
\end{table}

To gain deeper insights into whether current SLMs benefit from in-context learning, we conduct additional experiments on GLM-4-Voice. Specifically, we evaluate three mathematical domains in VoxEval using varying numbers of in-context learning examples, assessing GLM-4-Voice performance both with and without Chain-of-Thought (CoT) prompts. We use the three subject data in Appendix \ref{sec:comparisonspeechinteractionmodels}. Our analysis of the results presented in Table \ref{tab:morecotresults} reveals three key findings.
\begin{itemize}[left=0pt]
    \item \textbf{In-context examples improve performance for both CoT and non-CoT approaches.} The 1-shot configurations consistently outperform 0-shot baselines across both CoT and non-CoT settings, demonstrating that SLMs can effectively leverage contextual examples to enhance their mathematical reasoning capabilities.
    \item \textbf{CoT effectiveness varies inversely with problem complexity.} While CoT prompting consistently outperforms non-CoT approaches in elementary mathematics, this advantage diminishes for more advanced topics (high school and college mathematics). This pattern suggests that GLM-4-Voice can effectively utilize reasoning paths provided in CoT prompts for simpler problems, indicating the presence of reasoning capabilities in SLMs. However, these benefits do not extend to more complex mathematical domains.
    \item \textbf{GLM-4-Voice exhibits a trade-off between instruction-following and reasoning abilities.} Notably, 1-shot CoT underperforms 1-shot non-CoT on more challenging mathematical problems. This phenomenon suggests that in non-CoT settings, SLMs may rely on simple guessing rather than explicit reasoning. Conversely, when CoT prompts guide the model toward explicit reasoning on problems beyond its reasoning capacity, the model may become overly focused on the reasoning process at the expense of proper answer formatting and selection.
\end{itemize}

\section{Statistical Test Results across Different Input Audio Conditions}
\begin{table}[t]
  \centering
\scalebox{1.0}{
  \begin{tabular}{lcc} 
    \toprule
    \textbf{Models}  & \textbf{$\chi^2$(10)} & \textbf{p-value} \\
    \midrule
    GLM-4-Voice       & 121.53 & 2.47e-21  \\
    Moshi       & 180.54 & 1.81e-33  \\
    SPIRIT-LM       & 41.09 & 1.09e-05  \\
    TWIST       & 254.76 & 5.40e-49  \\
    SpeechGPT       & 7.97 & 0.63  \\
    \bottomrule
  \end{tabular}
}
  \caption{\label{tab:friedmantestresults}
    Friedman Test results for the five SLMs evaluated in this paper.
  }
\end{table}

\begin{figure*}[t]
  \centering
  \begin{subfigure}[b]{0.45\textwidth}
      \includegraphics[width=\textwidth]{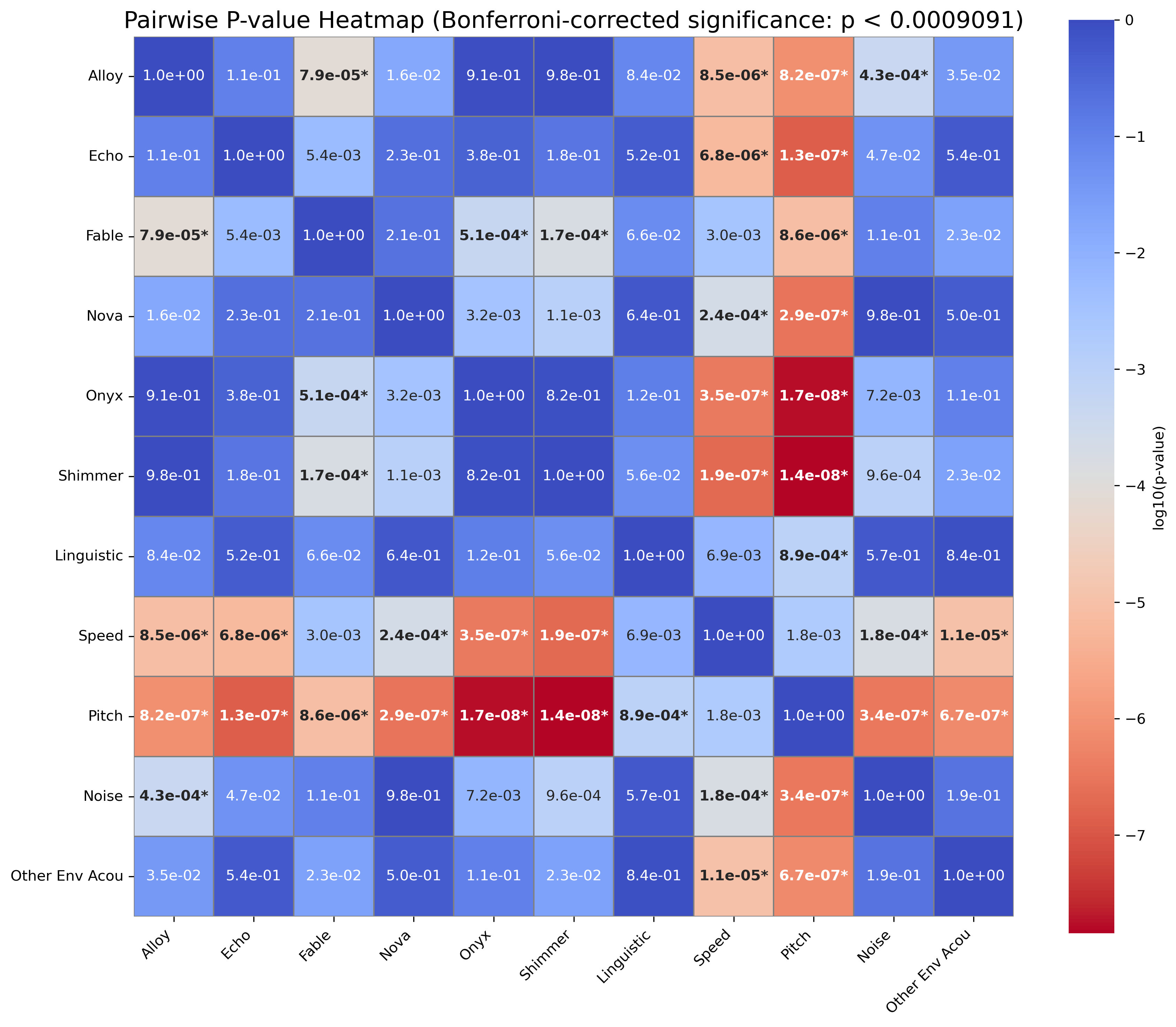}
      \caption{GLM-4-Voice}
  \end{subfigure}
  \hfill
  \begin{subfigure}[b]{0.45\textwidth}
      \includegraphics[width=\textwidth]{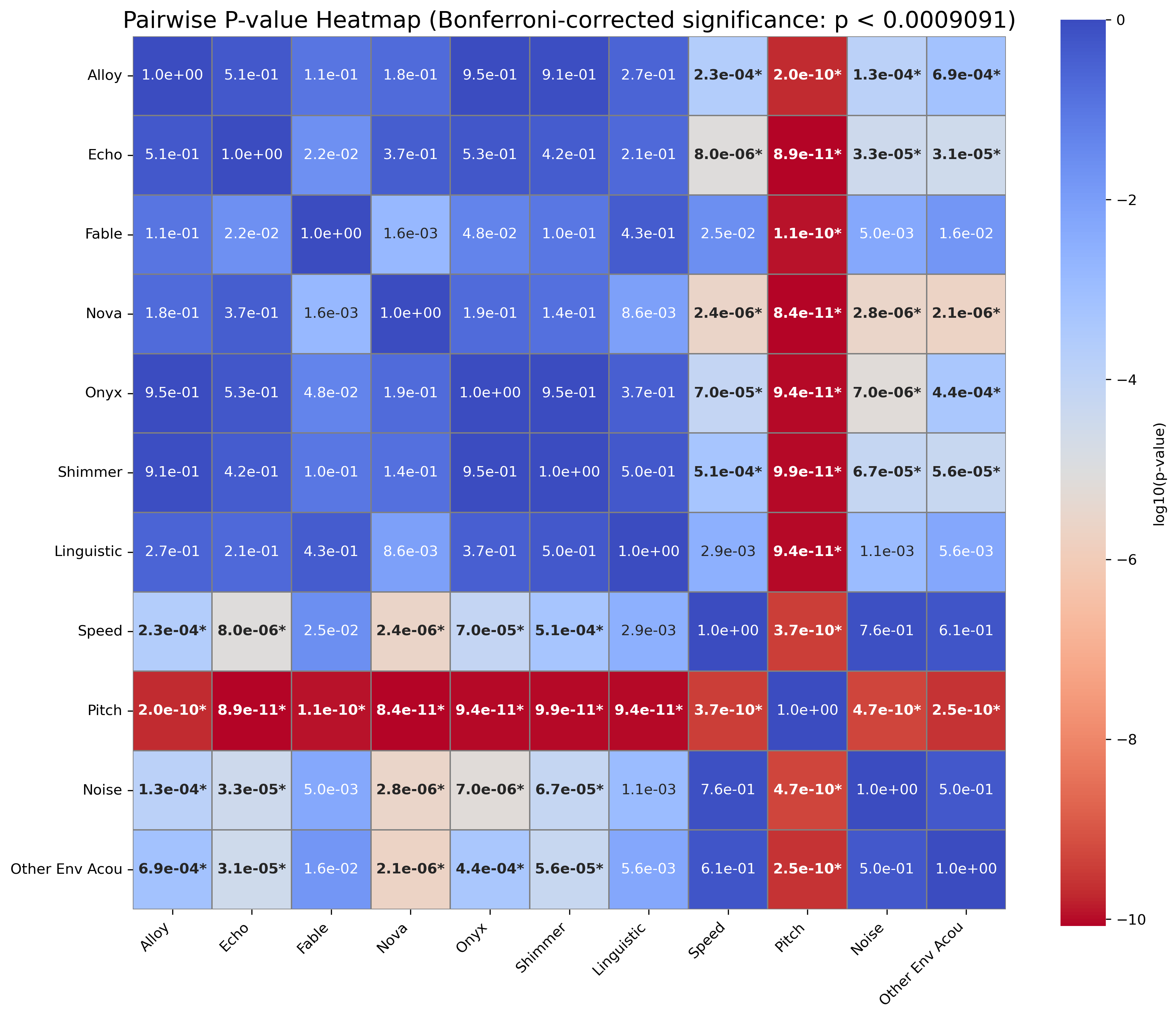}
      \caption{Moshi}
  \end{subfigure}
  
  \vspace{1em} 
  
  \begin{subfigure}[b]{0.45\textwidth}
      \includegraphics[width=\textwidth]{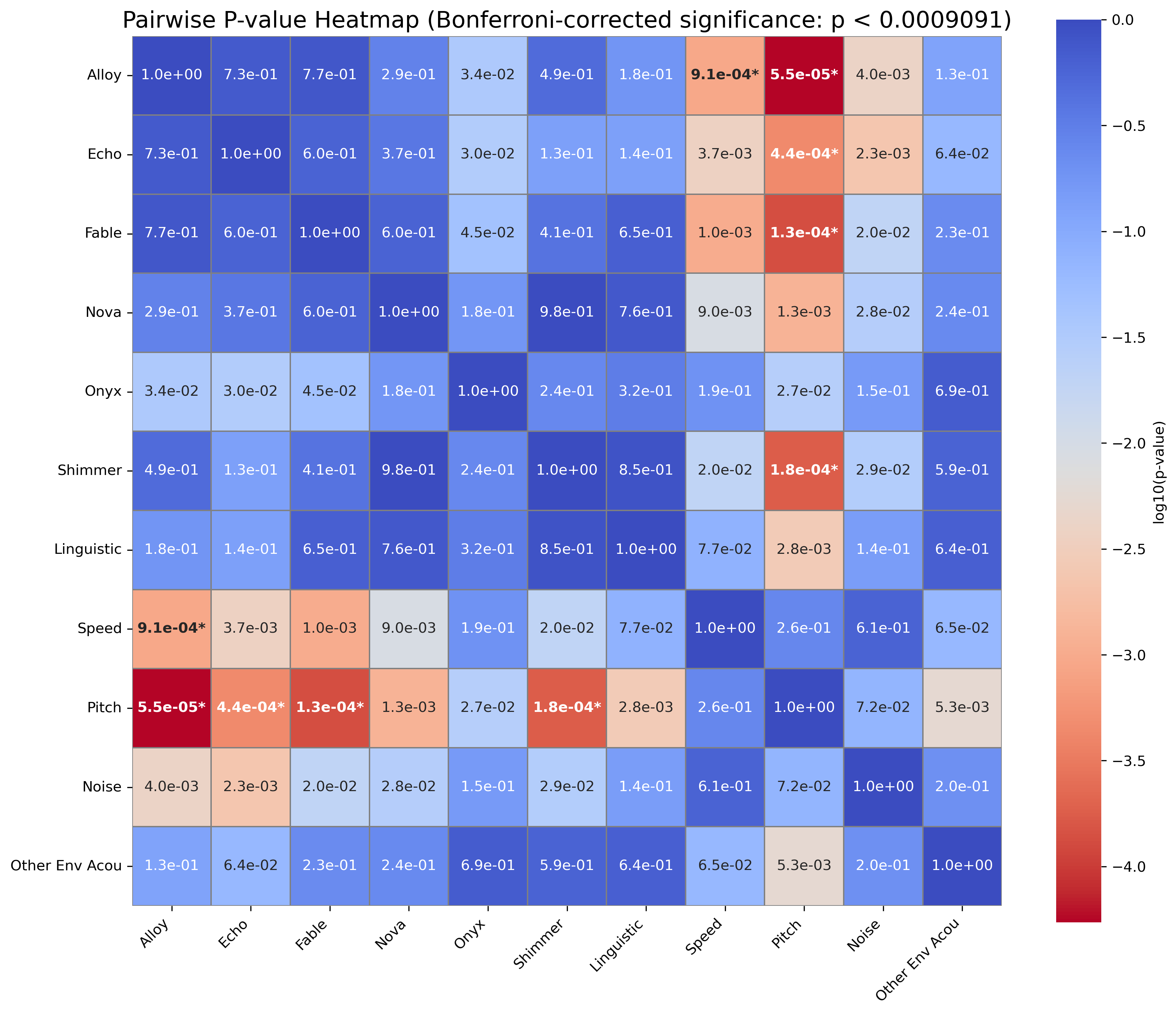}
      \caption{SPIRIT-LM}
  \end{subfigure}
  \hfill
  \begin{subfigure}[b]{0.45\textwidth}
      \includegraphics[width=\textwidth]{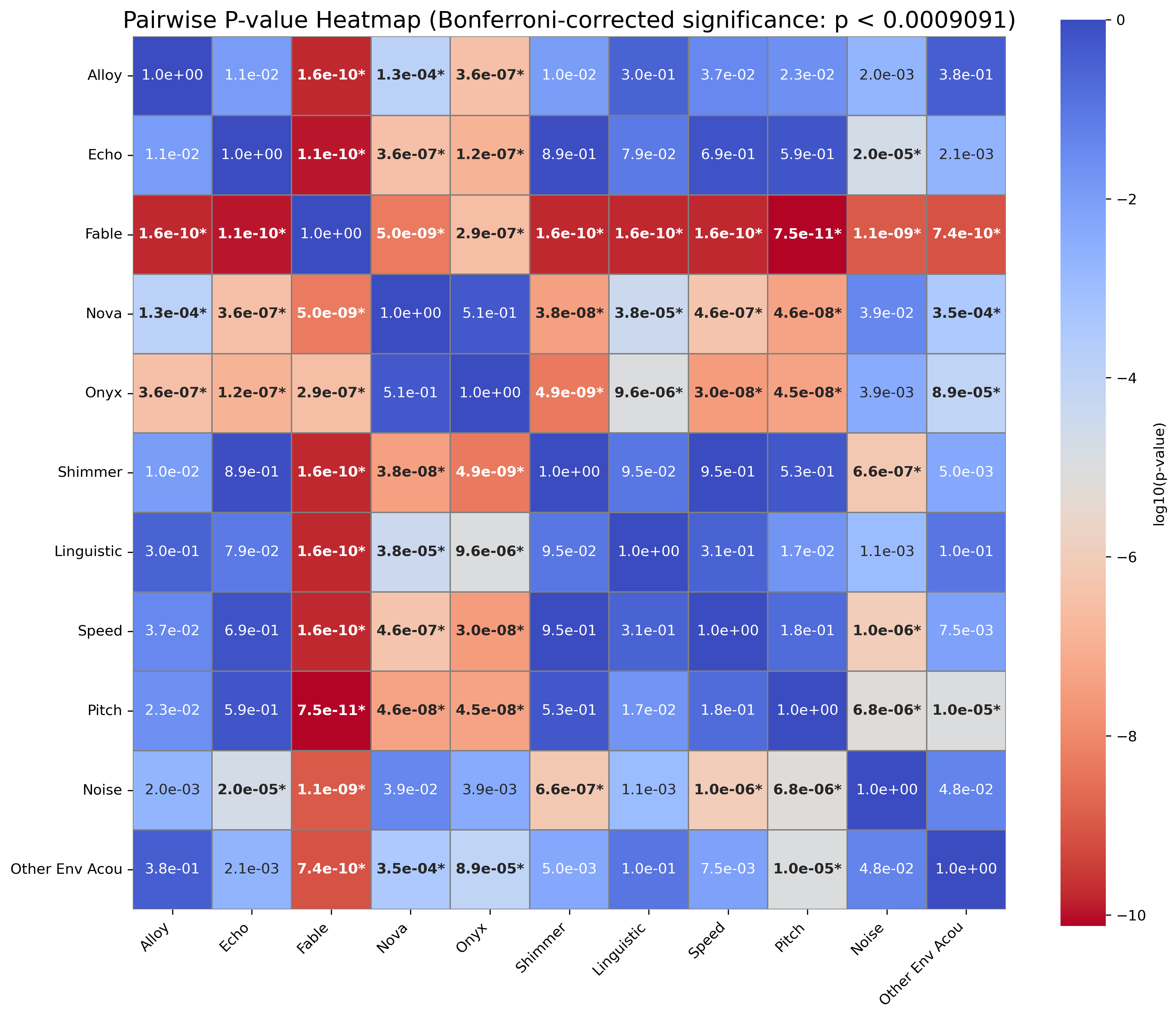}
      \caption{TWIST}
  \end{subfigure}
  \caption{Wilcoxon signed-rank test results for the four SLMs showing statistically significant differences in the Friedman test.}
  \label{fig:statsheatmaps}
\end{figure*}

To better understand how various input audio conditions influence the performance of different SLMs, we conduct statistical analyses on the models' responses across the 11 distinct audio conditions (see Table \ref{tab:mainresult}) for each SLM. Specifically, we collect the 56 subject accuracy scores of each SLM under each audio condition and apply the Friedman Test to analyze all conditions, followed by Wilcoxon Signed-Rank Tests for pairwise comparisons between conditions. A Type I error $\alpha$ of 5\% is used in the statistical analyses.

The Friedman Test results, summarized in Table \ref{tab:friedmantestresults}, reveal that input audio conditions significantly affect the performance of all SLMs except SpeechGPT, which consistently exhibits near-zero accuracy across all conditions. To further investigate these effects, we conduct pairwise Wilcoxon Signed-Rank Tests with Bonferroni correction on the remaining four models. The corrected significance level is calculated as:
\begin{equation}
\alpha_{corrected} = \frac{\alpha}{n} = \frac{0.05}{55} = 0.000909,
\end{equation}
where
\begin{equation}
n = \binom{11}{2} = 55
\end{equation}
represents the total number of pairwise comparisons among the 11 audio conditions. The resulting p-values are visualized as heatmaps in Figure \ref{fig:statsheatmaps}, with statistically significant values highlighted in bold. The heatmaps demonstrate that the impact of input audio conditions varies across different SLMs. Notably, GLM-4-Voice, Moshi, and SPIRIT-LM exhibit significant performance differences under the "Speed" and "Pitch" conditions, suggesting that these models are particularly sensitive to variations in speaking rate and pitch. This finding underscores the importance of robust handling of diverse user speech patterns in SLM design.

\begin{algorithm}[H]
\caption{Procedure for Generating Box Plots of Performance Differences}
\label{alg:boxplot}
\small
\begin{algorithmic}[1]
\STATE \textbf{Input:} VoxEval dataset with $N=56$ subjects, input audio condition settings $\{S_1, S_2, \dots, S_k\}$, and corresponding groups of results for each setting.
\STATE \textbf{Output:} Box plots visualizing performance differences for each setting.

\FOR{each input audio condition setting $S_i$}
    \STATE Extract all result groups $\{G_1, G_2, \dots, G_m\}$ corresponding to $S_i$.
    \STATE Initialize an empty list $D_i = []$ to store difference scores for $S_i$.
    
    \FOR{each subject $n \in \{1, 2, \dots, N\}$}
        \STATE Collect scores $\{s_{n,1}, s_{n,2}, \dots, s_{n,m}\}$ for subject $n$ across all $m$ groups in $S_i$.
        \STATE Compute the maximum absolute score difference:
        \[
        d_n = \max_{j,k} \big| s_{n,j} - s_{n,k} \big| \quad \text{for } j, k \in \{1, 2, \dots, m\}.
        \]
        \STATE Append $d_n$ to $D_i$: $D_i \gets D_i \cup \{d_n\}$.
    \ENDFOR

    \STATE Use $D_i = \{d_1, d_2, \dots, d_{56}\}$ to generate a box plot for $S_i$.
\ENDFOR

\STATE \textbf{Output:} Box plots for all input audio condition settings $\{S_1, S_2, \dots, S_k\}$.

\end{algorithmic}
\end{algorithm}

\section{Explanation for Various Speech Recording Conditions}
\label{apdx:recordingconditions}
Table \ref{tab:audioconditions} provides the input audio conditions categorized under the ``other environment acoustics" and the transformation settings in the \textit{audioaugmentation} package.

\begin{table}[t]
  \centering
\scalebox{1.0}{
  \begin{tabular}{ll}
    \toprule
    \textbf{Noise Type}  & \textbf{Settings} \\
    \midrule
    \multirow{2}{*}{Gaussian Noise} & min\_snr\_db=5.0 \\
                                    & max\_snr\_db=20.0 \\
    \midrule
    \multirow{2}{*}{Colored Noise}  & min\_snr\_db=5.0 \\
                                    & max\_snr\_db=20.0 \\
    \midrule
    \multirow{2}{*}{Background Music} & min\_snr\_db=3.0 \\
                                      & max\_snr\_db=10.0 \\
    \midrule
    \multirow{2}{*}{\begin{tabular}[c]{@{}l@{}}Short Interruption\\Noise\end{tabular}} & min\_snr\_db=3.0 \\
                                      & max\_snr\_db=10.0 \\
    \bottomrule
  \end{tabular}
}
  \caption{\label{tab:noiseparameters}
    Transformation settings for the various types of noise used.
  }
\end{table}

\begin{table*}[t]
  \centering
  \rowcolors{1}{white}{lightblue}
  \small 
  \begin{tabular}{lp{7.4cm}p{4cm}} 
    \toprule
    \textbf{Audio Conditions}  & \textbf{Explanation}  & \textbf{Settings} \\
    \midrule
    Aliasing       & Aliasing occurs when a signal is sampled at a rate lower than twice its highest frequency (violating the Nyquist theorem), causing higher frequencies to fold back and appear as lower, incorrect frequencies in the sampled signal. & min\_sample\_rate=4000, max\_sample\_rate=6000   \\
    Room Impulse Response       & Room impulse response refers to the acoustic fingerprint of a space, capturing how sound propagates, reflects, and decays in that environment. & A bedroom impulse response is used.   \\
    Low Pass Filter       & A low-pass filter in audio processing allows low-frequency sounds to pass through while attenuating (reducing) higher frequencies, helping to remove noise or create a smoother sound. & min\_cutoff\_freq=5000.0, max\_cutoff\_freq=5000.0   \\
    Band Pass Filter       & A band-pass filter in audio processing allows mid-frequency sounds to pass through & min\_center\_freq=2000.0, max\_center\_freq=2000.0   \\
    High Pass Filter     & A high-pass filter in audio processing allows high-frequency sounds to pass through  & min\_cutoff\_freq=1000.0, max\_cutoff\_freq=1000.0   \\
    Bitcrush      & Bitcrush reduces the resolution (bit depth) and/or sampling rate of a sound, creating a lo-fi, distorted, or ``crunchy" effect by introducing digital artifacts and aliasing. & min\_bit\_depth=5, max\_bit\_depth=6   \\
    Gain       & Gain refers to the adjustment of the amplitude or volume of an audio signal. & min\_gain\_db=-12.0, max\_gain\_db=12.0   \\
    Clipping Distortion       & Clipping distortion occurs when an audio signal exceeds the maximum amplitude that a system can handle, causing the waveform to be ``clipped" at the peaks and resulting in a harsh, distorted sound. & min\_percentile\_threshold=20, max\_percentile\_threshold=20   \\
    Seven Band Parameter EQ       & A Seven-Band Parametric EQ is an equalizer that allows precise control over seven specific frequency bands, enabling adjustments to the gain, bandwidth (Q), and center frequency of each band to shape the tonal balance or correct issues in the audio signal. & min\_gain\_db=-12.0, max\_gain\_db=12.0   \\
    \bottomrule
  \end{tabular}
  \caption{\label{tab:audioconditions}
    Explanations and transformation settings categorized under ``other environment acoustics."
  }
\end{table*}

\section{Additional Experimental Settings}
For models used in our evaluation, SpeechGPT, TWIST, SPIRIT-LM, and Moshi contain 7B parameters, and GLM-4-Voice contains 9B parameters. We use the base version of SPIRIT-LM. 
For all the models, we use the default hyperparameters provided in their code implementations. Each experiment is performed once.

\section{Case Study for SLM Responses}
\label{apdx:SLMresponses}
Table \ref{tab:casestudy} presents example SLM responses to questions from VoxEval as well as our comments. It highlights the most characteristic answer style for each SLM. We randomly choose responses from each SLM under the \textit{Alloy} setting while ensuring that the selected answer style is the most representative of that particular SLM.

\begin{table*}[t]
  \centering
  \rowcolors{1}{white}{lightblue}
  \small 
  \begin{tabular}{lp{5cm}p{8cm}} 
    \toprule
    \textbf{SLMs}  & \textbf{Example Response} & \textbf{Comment} \\
    \midrule
    SpeechGPT       & I bet. & SpeechGPT outputs random words for the most of time. \\
    TWIST       & A, C, A, and E plus, that's where H, H, and A, F, and E, D, F, A, A, & TWIST sometimes gives an answer but cannot form an interpretable sentence for the most of time. \\
    SPIRIT-LM       &  The correct answer is A. Statement 1. K is a normal subgroup of X. X is a normal subgroup of X. Please choose the answer of the question from Options A, B, C, or D. & SPIRIT-LM is able to follow the required format and output an answer, but it does not stop after giving an answer. \\
    Moshi       &  I think it's D D & Moshi is able to interpret the question and output an answer, but it does not follow the format of ``The correct answer is x.". \\
    GLM-4-Voice       &  The correct answer is B. Multiplication is not associative. & GLM-4-Voice can respond in the specified format and provide an explanation afterward, though the explanation may occasionally be inaccurate. \\
    \bottomrule
  \end{tabular}
  \caption{\label{tab:casestudy}
    Examples of each SLM's response and our comments.
  }
\end{table*}

\section{String Matching vs. LLM Extraction}
In Table \ref{tab:mathresult}, we notice that in most cases, the accuracy of using an LLM to extract the final answer is lower. This is primarily due to edge cases. For instance, when the SLM generates an output like 
\begin{lstlisting}[style=mystyle]
"is actually to where h or 0 0 1 times 2 squared. Option D, to where v is a positive, moving by 1 times 2 squared. Statement 1, because of the path grow entertained by t is", 
\end{lstlisting}
the string matching algorithm identifies D as the final answer. However, the SLM does not explicitly state that D is the correct answer. In such situations, the LLM produces ``None!" as the output. If D happens to be the correct answer, string matching would classify it as correct, whereas the LLM would mark it as incorrect. The full prompt used for extracting the final answer with the LLM is provided in Figure \ref{fig:llmanswerextraction}.

\begin{figure*}[ht]
    \centering
    \begin{lstlisting}[style=mystyle]
Please convert the following statement into a human-readable format. For example, please convert Arabic numerals into words, and convert other elements like operation symbols, mathematical units, abbreviations and any special characters into their corresponding words. If you see a sequence of underscores, that means we need to mask it out (see the examples below). Some statements may not contain special symbols, so you will need to determine if any conversion is necessary. Ensure that you output the entire statement without omitting any part of it. Do not include any additional text beyond the converted statement.

Below are some general examples:
# Statement
25cm
# Answer
Twenty-five centimeters
# Statement
The period of a 10-Hz wave is 1/10s.
# Answer
The period of a ten-hertz wave is one-tenth seconds.
# Statement
Find the value of $4 \div 2 \cdot (2 + 8) - 4$.
# Answer
Find the value of four divided by two times the sum of two and eight, minus four.
# Statement
dim (U) + dim (W) = n^2
# Answer
The dimension of U plus the dimension of W equals n squared.
# Statement
The _______ developed one of the earliest kingdoms in South America, reaching its height at around _________.
# Answer
The, masked, developed one of the earliest kingdoms in South America, reaching its height at around, masked.
# Statement
I have some fruits: I. Apple. II. Pear. III. Banana.
# Answer
I have some fruits: One, apple. Two, pear. Three, banana.

Now, this is the statement I want you to convert:
# Statement
{statement}
# Answer 
    \end{lstlisting}
    \caption{The prompt for GPT-4o to convert questions with math expressions.}
    \label{fig:questionprompt}
\end{figure*}

\begin{figure*}[ht]
    \centering
    \begin{lstlisting}[style=mystyle]
Please convert the following statement into a human-readable format. For example, please convert Arabic numerals into words, and convert other elements like operation symbols, mathematical units, abbreviations and any special characters into their corresponding words. If you see a sequence of underscores, that means we need to mask it out (see the examples below). Some statements may not contain special symbols, so you will need to determine if any conversion is necessary. Ensure that you output the entire statement without omitting any part of it. Do not include any additional text beyond the converted statement.

Below are some general examples:
# Statement
25cm
# Answer
Twenty-five centimeters
# Statement
The period of a 10-Hz wave is 1/10s.
# Answer
The period of a ten-hertz wave is one-tenth seconds.
# Statement
Find the value of $4 \div 2 \cdot (2 + 8) - 4$.
# Answer
Find the value of four divided by two times the sum of two and eight, minus four.
# Statement
dim (U) + dim (W) = n^2
# Answer
The dimension of U plus the dimension of W equals n squared.
# Statement
The _______ developed one of the earliest kingdoms in South America, reaching its height at around _________.
# Answer
The, masked, developed one of the earliest kingdoms in South America, reaching its height at around, masked.
# Statement
I have some fruits: I. Apple. II. Pear. III. Banana.
# Answer
I have some fruits: One, apple. Two, pear. Three, banana.

Additionally, when converting the "choice" within an MCQ, please convert the choice based on the contexts presented in the question. Some examples:
# Question Context
A rod measures 1.00 m in its rest system. How fast must an observer move parallel to the rod to measure its length to be 0.80 m?
# Choice
0.80c
# Answer
Zero point eight times the speed of light.
# Question Context
A spring of force constant k is stretched a certain distance. It takes twice as much work to stretch a second spring by half this distance. The force constant of the second spring is
# Choice
4k
# Answer
Four k.
# Question Context
4k times 2 is
# Choice
8k
# Answer
Eight thousand.

Now, below is the "choice" I want you to convert:
# Question Context
{question}
# Choice
{choice}
# Answer
    \end{lstlisting}
    \caption{The prompt for GPT-4o to convert answer choices with math expressions.}
    \label{fig:answerchoiceprompt}
\end{figure*}

\begin{figure*}[ht]
    \centering
    \begin{lstlisting}[style=mystyle]
{SLM output}

# Instruction
Above is the answer provided by an AI model for a Multiple Choice Question with four answer choices (A, B, C, or D). Based on the above text, extract the final answer from it. If you can find the answer from the above text, only output the answer choice. Do not include anything else in your output. For example, a possible output is "C". If you cannot find the answer, only output "None!".
    \end{lstlisting}
    \caption{The prompt for GPT-4o to extract the final answers from SLM outputs.}
    \label{fig:llmanswerextraction}
\end{figure*}

\end{document}